\documentclass[runningheads]{llncs}
\usepackage[T1]{fontenc}
\usepackage{colortbl}
\usepackage{array,multirow,graphicx}
\usepackage{capt-of}
\usepackage{color}
\usepackage{algorithm}
\usepackage[noend]{algpseudocode}
\usepackage{hyperref}
\usepackage{tikz}
\usepackage{comment,nicefrac}
\usepackage{graphicx}
\usepackage{capt-of}
\usepackage{bbm}
\usepackage{amsmath,amssymb}
\usepackage{booktabs}
\usepackage{epigraph,booktabs}
\usepackage{rotating}
\usepackage{tabularx}
\usepackage{adjustbox}
\usepackage{pifont}       % \ding{xx}
\usepackage{bbding}       % \Checkmark,\XSolid,... (需要和pifont宏包共同使用)
\usepackage{fontawesome}  % \faCheck,\faTimes

\newcommand{\etal}{\textit{et al.}}

\usepackage[section]{placeins}

\begin{document}

\title{Medical Image Synthesis via \\ Fine-Grained Image-Text Alignment and \\Anatomy-Pathology Prompting}
\titlerunning{Medical Image Synthesis}
\authorrunning{Chen \etal}
% \author{}%Paper 3619}

\author{Wenting Chen$^1$, Pengyu Wang$^2$, Hui Ren$^3$, Lichao Sun$^4$, Quanzheng Li$^3$, Yixuan Yuan$^{2*}$, and Xiang Li$^3$\thanks{Corresponding authors: Yixuan Yuan (\href{mailto:yxyuan@ee.cuhk.edu.hk}{yxyuan@ee.cuhk.edu.hk}), Xiang Li (\href{mailto:xli60@mgh.harvard.edu}{xli60@mgh.harvard.edu})}
% $^1$City University of Hong Kong~$^2$The Chinese University of Hong Kong\\ [0.5mm]
% $^4$Massachusetts General Hospital and Harvard Medical School\\ [0.5mm]
% $^3$Lehigh University\\ [0.5mm]
}
\institute{$^1$City University of Hong Kong~$^2$The Chinese University of Hong Kong\\ [0.5mm]
$^3$Massachusetts General Hospital and Harvard Medical School\\ [0.5mm]
$^4$Lehigh University\\ [0.5mm]}

% Pengyu Wang (The Chinese University of Hong Kong) <pengyuwang@cuhk.edu.hk> 
% Hui Ren (HMS/MGH) <hren2@mgh.harvard.edu> 
% Lichao Sun (Lehigh University) <james.lichao.sun@gmail.com> 
% Quanzheng Li (Massachusetts General Hospital and Harvard Medical School) <li.quanzheng@mgh.harvard.edu> 
% Yixuan Yuan (Chinese University of Hong Kong) <yxyuan@ee.cuhk.edu.hk> 
% Xiang Li (Massachusetts General Hospital and Harvard Medical School) <xli60@mgh.harvard.edu> 
\maketitle              % typeset the header of the contribution
%Morever, the information imbalance and inter-modal gap between medical reports and images pose a challenge to 
\begin{abstract}
Data scarcity and privacy concerns limit the availability of high-quality medical images for public use, which can be mitigated through medical image synthesis. However, current medical image synthesis methods often struggle to accurately capture the complexity of detailed anatomical structures and pathological conditions. To address these challenges, we propose a novel medical image synthesis model that leverages fine-grained image-text alignment and anatomy-pathology prompts to generate highly detailed and accurate synthetic medical images. Our method integrates advanced natural language processing techniques with image generative modeling, enabling precise alignment between descriptive text prompts and the synthesized images' anatomical and pathological details. The proposed approach consists of two key components: an anatomy-pathology prompting module and a fine-grained alignment-based synthesis module. The anatomy-pathology prompting module automatically generates descriptive prompts for high-quality medical images. To further synthesize high-quality medical images from the generated prompts, the fine-grained alignment-based synthesis module pre-defines a visual codebook for the radiology dataset and performs fine-grained alignment between the codebook and generated prompts to obtain key patches as visual clues, facilitating accurate image synthesis. We validate the superiority of our method through experiments on public chest X-ray datasets and demonstrate that our synthetic images preserve accurate semantic information, making them valuable for various medical applications.
\vspace{0.5cm}
\end{abstract}
\section{Introduction}

In the medical field, high-quality medical images are scarce and difficult to access due to data privacy concerns and the labor-intensive process of collecting such data~\cite{el2022overcome}. This scarcity of medical images can hinder the development and training of artificial intelligence (AI) models for various medical applications, such as diagnosis, segmentation, and abnormality classification. One solution to overcome this challenge is to use medical image synthesis techniques to generate synthetic data that can replace or supplement real medical images.

Several chest X-ray generation methods have been investigated to mitigate these issues, which can be categorized into three main groups: generative adversarial networks (GAN) based~\cite{madani2018chest,zhang2019skrgan,karbhari2021generation}, diffusion based~\cite{chambon2022adapting,chambon2022roentgen}, and transformer based~\cite{lee2023unified,lee2023llm} methods. Madani~\etal~\cite{madani2018chest} and Zhang~\etal~\cite{zhang2019skrgan} utilize unconditional GANs to synthesize medical images as a form of data augmentation to improve segmentation and abnormality classification performance. To leverage medical reports, some diffusion-based methods~\cite{chambon2022adapting,chambon2022roentgen} take the impression section of medical reports and random Gaussian noise as input for chest X-ray generation, ignoring the finding section that includes more detailed descriptions. To consider more details in medical reports, several transformer-based methods~\cite{lee2023unified,lee2023llm} take both finding and impression sections of medical reports as input to synthesize chest X-rays. However, current methods generate medical images based on the given ground-truth report from the dataset, which may not fully describe all the details of the medical image. In fact, medical images contain different anatomical structures (lobe, heart, and mediastinal) and pathological conditions (opacity, effusion, and consolidation), which are important for clinical diagnosis. As a result, the generated medical images often lack this detailed information. Thus, there is a need for a medical image synthesis method that can generate high-quality medical images with detailed anatomical and pathological descriptions.

% ~\cite{yan2023attributed}
% Another challenge for current medical image synthesis methods is the huge inter-modal gap. Medical images, comprising thousands of pixels, visualize rich textures and colors, while medical reports consist of a few sentences to summarize the findings and impressions of the medical images. As a result, there is a great imbalance between their amount of information, leading to a large inter-modal gap between medical reports and images. Thus, it is necessary to mitigate the information imbalance and minimize the inter-modal gap.

Another significant challenge for current medical image synthesis methods is the substantial inter-modal gap between medical images and reports. Medical images, comprising thousands of pixels, visualize rich textures and colors, while medical reports consist of only a few sentences to summarize the findings and impressions of the medical images. This disparity leads to a great imbalance in the amount of information contained in each modality, resulting in a large inter-modal gap between medical reports and images~\cite{henning2017estimating}. As a result, the generated medical images may not accurately reflect the content of the corresponding medical reports, as the synthesis models struggle to bridge this information gap. Furthermore, the limited information provided in the medical reports may not be sufficient to guide the synthesis of highly detailed and accurate medical images, which are crucial for clinical diagnosis and decision-making. Thus, it is necessary to develop techniques that can effectively mitigate the information imbalance and minimize the inter-modal gap between medical reports and images. By doing so, the synthesized medical images can better capture the detailed anatomical structures and pathological conditions described in the medical reports, leading to more reliable and informative synthetic data for various medical applications.

To address these issues, we propose a novel medical image synthesis model that leverages the capabilities of fine-grained image-text alignment and anatomy-pathology prompts to generate highly detailed and accurate synthetic medical images. Our approach consists of two key components: an \textbf{anatomy-pathology prompting} and a \textbf{fine-grained alignment based synthesis module}. The \textbf{anatomy-pathology prompting} aims to automatically generate descriptive reports for high-quality medical images. It first constructs the anatomy and pathology vocabularies from radiology reports under the guidance of radiologists, and then employs GPT-4 to write reports based on the given vocabularies. This ensures that the generated reports contain comprehensive and accurate descriptions of the anatomical structures and pathological conditions present in the medical images. To further synthesize high-quality medical images from the generated reports, we introduce a \textbf{fine-grained alignment based synthesis module}. This module pre-defines a visual codebook containing multiple patches commonly observed in the radiology dataset and performs fine-grained alignment between the generated reports and the visual codebook. Through this alignment, the module extracts the most matched keypatches that provide visual clues for the large language model (LLM) during the synthesis process. The LLM takes the generated reports, keypatches, and instructions as input and outputs visual tokens, which are then decoded by a VQ-GAN decoder to produce the final synthetic medical images. We conduct extensive experiments on publicly available chest X-ray (CXR) datasets to validate the superiority of our method compared to existing approaches. Furthermore, we perform semantic analysis on both real and synthetic images to demonstrate that our synthetic images preserve accurate semantic information, including anatomical structures and pathological conditions, making them valuable for various medical applications.

\begin{figure}[t]
  \centering
   \includegraphics[width=\linewidth]{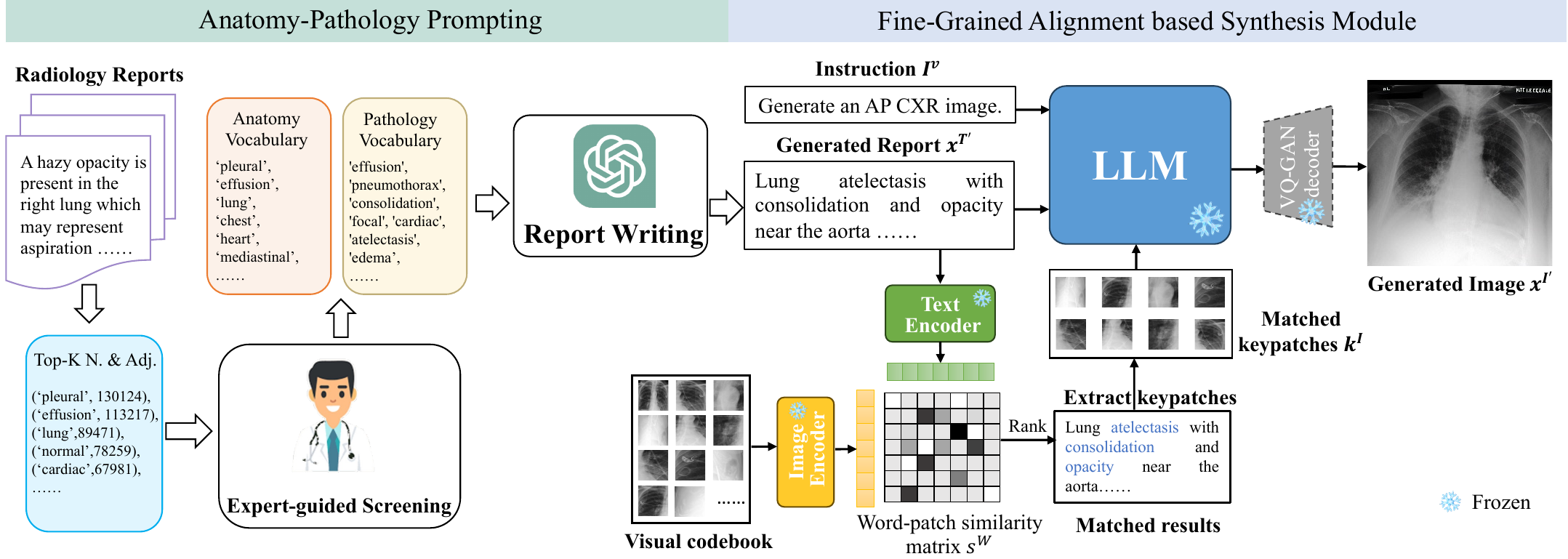}
   \vspace{0.2cm}
   \caption{The overview of the proposed method. It consists of an anatomy-pathology prompting module to generate descriptive reports with given anatomy and pathology words, and a fine-grained alignment based synthesis module using fine-grained image-text alignment to facilitate image generation.}
   \label{fig:overview}
\end{figure}

\newpage
\section{Method}
% In this section, we introduce our proposed medical image synthesis model, which includes an anatomy-pathology prompting module and a fine-grained alignment-based synthesis module.
\subsection{Anatomy-Pathology Prompting}
Since current methods struggle to synthesize medical images with complex anatomical structures (lobe, heart, and mediastinal) and pathological conditions (opacity, effusion, and consolidation), we introduce an anatomy-pathology prompting to automatically generate descriptive reports for high-quality medical image generation. This prompting module contains two main steps, including the design of anatomy and pathology vocabularies and prompts generation.

\begin{figure}[t]
  \centering
   \includegraphics[width=\linewidth]{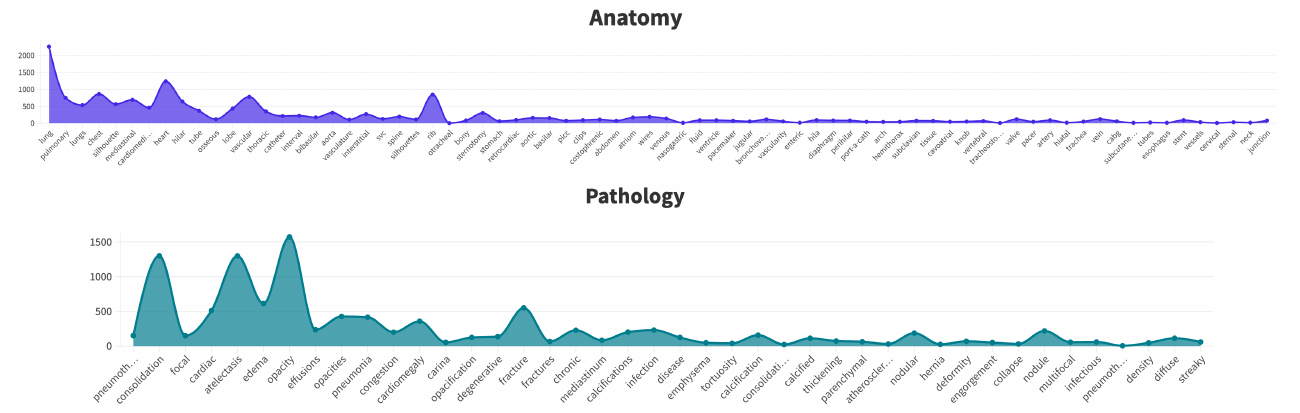}
   \caption{The word frequency of the anatomy and pathology vocabularies.}
   \label{fig:frequency}
\end{figure}

\noindent\textbf{Designing Anatomy and Pathology Vocabularies.} 
% As shown in Fig.~\ref{fig:overview}, we design the anatomy and pathology vocabularies that can extract instance-level anatomy and pathology words from reports and images. Considering that anatomy and pathology words belong to nouns and adjectives, we adopt a word filter to extract whole nouns and adjectives from the impression and finding parts of reports in the MIMIC-CXR dataset~\cite{johnson2019mimic}, and select the top-K nouns and adjectives according to their occurrence frequencies. Finally, we consult the expert’s guidance to manually remove the other non-medical nouns and adjectives that GPT-4 failed to screen out, and divide the screened words into the anatomy and pathology vocabularies according to their medical attributes.
As illustrated in Fig.~\ref{fig:overview}, we have developed anatomy and pathology vocabularies to extract instance-level anatomical and pathological terms from radiological reports and images. Recognizing that anatomical and pathological terms are typically nouns and adjectives, we employ a word filter to extract all nouns and adjectives from the impression and findings sections of reports in the MIMIC-CXR dataset~\cite{johnson2019mimic}. We then select the top-K nouns and adjectives based on their occurrence frequencies. Finally, under expert guidance, we manually remove any remaining non-medical nouns and adjectives that GPT-4 is unable to filter out, and categorize the screened words into anatomy and pathology vocabularies according to their medical attributes. The number of words in anatomy and pathology vocabularies is 75 and 44, respectively. We demonstrate the word frequency of the anatomy and pathology vocabularies, as shown in Fig.~\ref{fig:frequency}.

\noindent\textbf{Prompts Generation.} 
With the anatomy and pathology vocabularies, we employ GPT4 to automatically generate the medical reports. Specifically, we first provide the vocabularies to GPT4 and require it to randomly select $N$ and $M$ words from anatomy and pathology vocabularies, respectively, which can be combined as the findings. Then, these words are passed to GPT4 to write a report with reasonable findings for a chest X-ray image. To let GPT4 write reports as our requirement, we use the following instructions.

\vspace{8pt}
\noindent\fbox{\begin{minipage}{0.95\linewidth}
\scriptsize
% \tiny
\texttt{anatomy\_list = [`pleural', `lung', ......,`neck', `junction']\\
pathology\_list = [`effusion', `pneumothorax', ......, `diffuse', `streaky']\\
Here are two lists of anatomy and pathology for chest X-rays. Please write some findings that only include 2 words from the anatomy list and 2 from the pathology list, and do not write any negative sentences in the findings. These four words can be randomly selected from the two lists, respectively. Please ensure the findings are reasonable for a chest x-ray in real medical scenarios. The output should be in 50 words.
Here is an example:\\
anatomy\_list = [`heart', `diaphragm']\\
pathology\_list = [`effusion', `opacity']\\
Findings: Presence of opacity observed near the heart and diaphragm regions suggestive of effusion. \\
Please generate the output in the following format: \\
anatomy\_list = [`word1', `word2']\\
pathology\_list = [`word3', `word4']\\
Findings:}
\end{minipage}}
\vspace{10pt}

This instruction example requires GPT4 to use two words from anatomy and pathology vocabularies, respectively. Actually, we can use more than two words and set $N$ and $M$ for the number of words we used in anatomy and pathology vocabularies, respectively.
Then, we collect the anatomy-pathology prompts generated by GPT4, where each prompt contains an anatomy word list (e.g. \texttt{[`heart', `diaphragm']}), a pathology word list (e.g. \texttt{[`effusion', `opacity']}), and a generated report (e.g. \texttt{Presence of opacity observed near the heart and diaphragm regions suggestive of effusion.}). With these generated anatomy-pathology prompts, we can provide the synthesis model descriptive reports with detailed anatomical structures and pathological conditions.

\subsection{Fine-Grained Alignment based Synthesis Module}
Since there is an information imbalance and the inter-modal gap between medical reports and images, we devise a fine-grained alignment based synthesis module to leverage the fine-grained image-text alignment to facilitate image generation. The fine-grained alignment between medical reports and visual codebook to obtain matched keypatches as a clue for image synthesis. This module includes three steps for medical image synthesis, i.e. visual codebook construction, keypatches extraction, and image synthesis.

\noindent\textbf{Visual Codebook Construction.} 
To construct a visual codebook, we first identify the most common patches in the training set images and designate them as keypatches. This process involves matching patches from CXR images with textual tokens from their corresponding medical reports. We select the top $\kappa_1$ CXR-report pairs that exhibit the highest report-to-CXR similarities, denoted as $s^T$. For each selected CXR-report pair, we calculate the maximum similarity between each textual token and the image patches, resulting in word-patch maximum similarity scores. The embeddings of textual tokens and image patches are extracted by the pre-trained text and encoders~\cite{chen2023fine}, respectively. These scores are then ranked, and the patches corresponding to the top $\kappa_2$ similarities are extracted and included in the visual codebook as keypatches. Each keypatch in the codebook consists of the patch itself and its associated features.
% Concretely, we first construct a visual codebook with the most common patches as keypatches. To collect the most common patches for images in the training set, we match the patches of CXR images with textual tokens of medical reports to obtain top $\kappa_1$ CXR-report pairs with the highest report-to-CXR similarities $s^T$. For each CXR-report pair, we compute the word-patch maximum similarity for each textual token, rank the word-patch maximum similarities, and extract the patches for top $\kappa_2$ similarities as keypatches in the visual codebook. Each keypatch includes its patch and the corresponding features.

\noindent\textbf{Keypatches Extraction.} 
% With the visual codebook, we match the features of keypatches in the visual codebook with textual tokens of the generated report to acquire the word-patch similarity matrix $s^W \in \mathbb{R}^{(\kappa_2\times \kappa_3)\times K}$, where $K$ is the number of textual tokens. To obtain keypatches related to the generated report, we rank the word-patch similarity along the dimension of keypatches, obtain the top $\kappa_4$ word-patch similarity for each textual token, and extract the features of corresponding keypatches $k^I$. 
With the visual codebook, we establish a correspondence between the features of keypatches and the textual tokens of the generated report. This is achieved by matching the features of each keypatch in the visual codebook with the textual tokens, resulting in the creation of a word-patch similarity matrix, denoted as $s^W \in \mathbb{R}^{(\kappa_1\times \kappa_2)\times K}$, where $K$ represents the total number of textual tokens in the report. To identify the keypatches that are most relevant to the generated report, we perform a ranking operation on the word-patch similarity matrix along the dimension of keypatches. For each textual token, we select the top $\kappa_3$ keypatches with the highest word-patch similarity scores. Finally, we extract the features of these selected keypatches, denoted as $k^I$, which serve as a compact representation of the visual information most closely associated with the textual content of the generated report.

\noindent\textbf{Image Synthesis.} 
% After obtaining the keypatches, we use a frozen VQ-GAN encoder $E$ to convert the matched keypatches $k^I$ into image tokens $E(k^I)$, and feed the pre-trained large language model (LLM)~\cite{chen2023fine} with the instruction, generated report, and image tokens of keypatches in the instruction-following format, as shown in Fig.~\ref{instruct_text2img}. Then, LLM predicts image tokens, decoded by the VQ-GAN into the generated CXR image $x^{I'}$.
After acquiring the keypatches, we employ a frozen VQ-GAN encoder~\cite{esser2021taming} $E$ to transform the matched keypatches $k^I$ into image tokens $E(k^I)$. These image tokens are then fed into a pre-trained large language model (LLM)\cite{chen2023fine} along with the instruction and the generated report. The input to the LLM follows an instruction-following format. By providing the LLM with the instruction, generated report, and image tokens of the keypatches, we enable the model to predict image tokens that correspond to the desired CXR image. Finally, the predicted image tokens are decoded using the VQ-GAN decoder, resulting in the generation of the CXR image $x^{I'}$. This process leverages the power of the pre-trained LLM to interpret the textual instruction and report, while utilizing the visual information encoded in the keypatches to guide the generation of a realistic and coherent CXR image.

%LLM predicts the image tokens and the VQ-GAN decoder decodes the image tokens into the corresponding generated CXR image $x^{I'}$. 
By adopting the fine-grained alignment based synthesis module, we can generate high-quality medical images with the detailed anatomical structures and pathological conditions described in the medical reports.

\section{Experiments and Results}
\subsection{Experiment Setting}
\noindent\textbf{Datasets.} In our experiments, we utilize two widely-used publicly available chest X-ray datasets: MIMIC-CXR~\cite{johnson2019mimic} and OpenI~\cite{demner2016preparing}. The MIMIC-CXR dataset is a large-scale dataset consisting of 473,057 images and 206,563 corresponding medical reports from 63,478 patients. We adhere to the official dataset splits, which allocate 368,960 samples for training, 2,991 for validation, and 5,159 for testing. On the other hand, the OpenI dataset is smaller in size, containing 3,684 report-image pairs. The dataset is divided into 2,912 samples for training and 772 for testing. %In our experiments, we use MIMIC-CXR dataset to obtain the anatomy and pathology vocabularies.

\begin{table*}[t]
	\centering
	\renewcommand{\arraystretch}{1.2}
    \setlength{\tabcolsep}{0.2cm}
	\caption{Comparison of report-to-CXR generation performance on the MIMIC-CXR and the OpenI datasets.}
	\begin{tabular}{c|c|c|c|c}
		\toprule
   & \multicolumn{2}{c|}{MIMIC-CXR}  & \multicolumn{2}{c}{OpenI}  \\\hline
Methods & FID $\downarrow$ & NIQE $\downarrow$ & FID $\downarrow$ & NIQE $\downarrow$\\\hline
Stable diffusion~\cite{rombach2022high} & 14.5194  & 5.7455  & 11.3305  & 5.7455 \\%\hline
Chambon~\textit{et al.}~\cite{chambon2022adapting} & 12.7408  & 4.4534  & 8.2887  & 4.4534 \\%\hline
RoentGen~\cite{chambon2022roentgen} & 13.1979  & 5.1286  & 6.5666  & 5.1286 \\%\hline
UniXGen~\cite{lee2023unified} & 14.0569  & 6.2759  & 7.5210  & 6.2759 \\%\hline
LLM-CXR~\cite{lee2023llm} & 11.9873  & 4.5876  & 5.9869  & 4.5876 \\\hline
\textbf{Ours} & 8.8213  & 4.1138  & 5.7455  & 4.1138 \\
		\bottomrule
	\end{tabular}
	\label{tab:img_quality}
 \end{table*}

\begin{figure}[t]
  \centering
   \includegraphics[width=\linewidth]{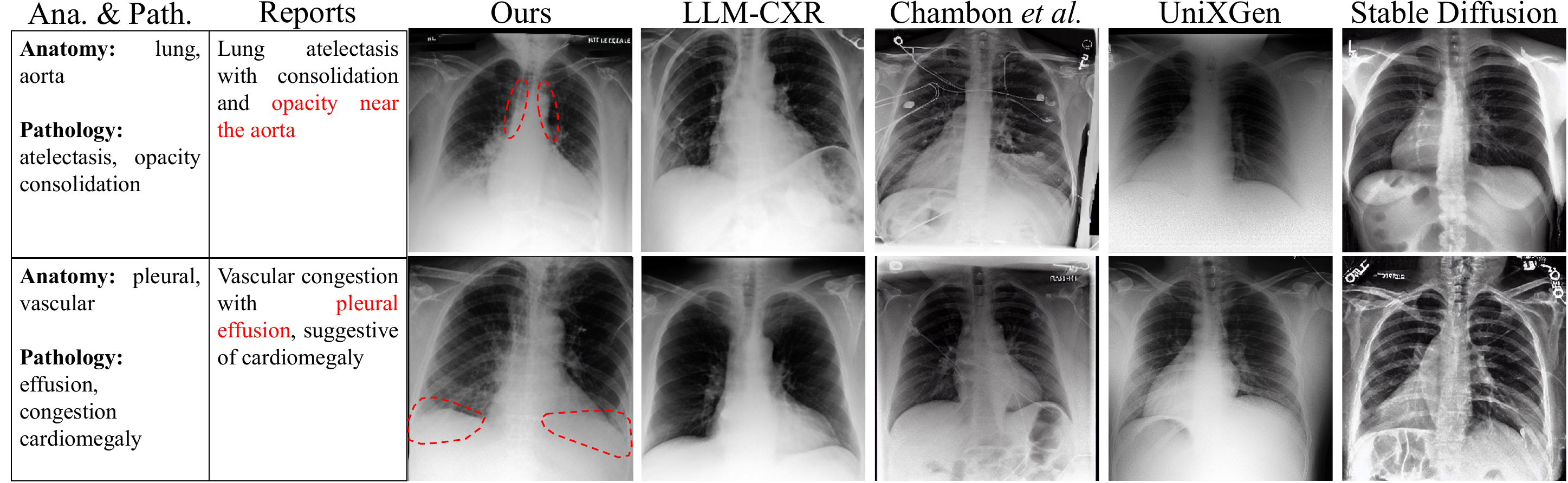}
   \caption{The generated chest X-ray images of the MIMIC-CXR dataset with highlighted regions.}
   \label{fig:report2cxr}
\end{figure}

\noindent\textbf{Implementation and Metrics.} We use the pre-trained image encoder, text encoder and LLM~\cite{chen2023fine} in the fine-grained alignment synthesis module. The pre-trained VQ-GAN model~\cite{esser2021taming} is adopted to encode image patches to image tokens, and decode the image tokens to images. All the models are frozen in the framework. 
To assess the image quality, we use the Fréchet inception distance (FID)~\cite{heusel2017gans} and Natural Image Quality Evaluator (NIQE)~\cite{mittal2012making}. The lower values indicate the better performance. %The source code will be released.
 
% \subsection{Ablation Study}
% \noindent\textbf{Effectiveness of Anatomy-Pathology Prompting.}

\begin{figure}[t]
  \centering
   \includegraphics[width=\linewidth]{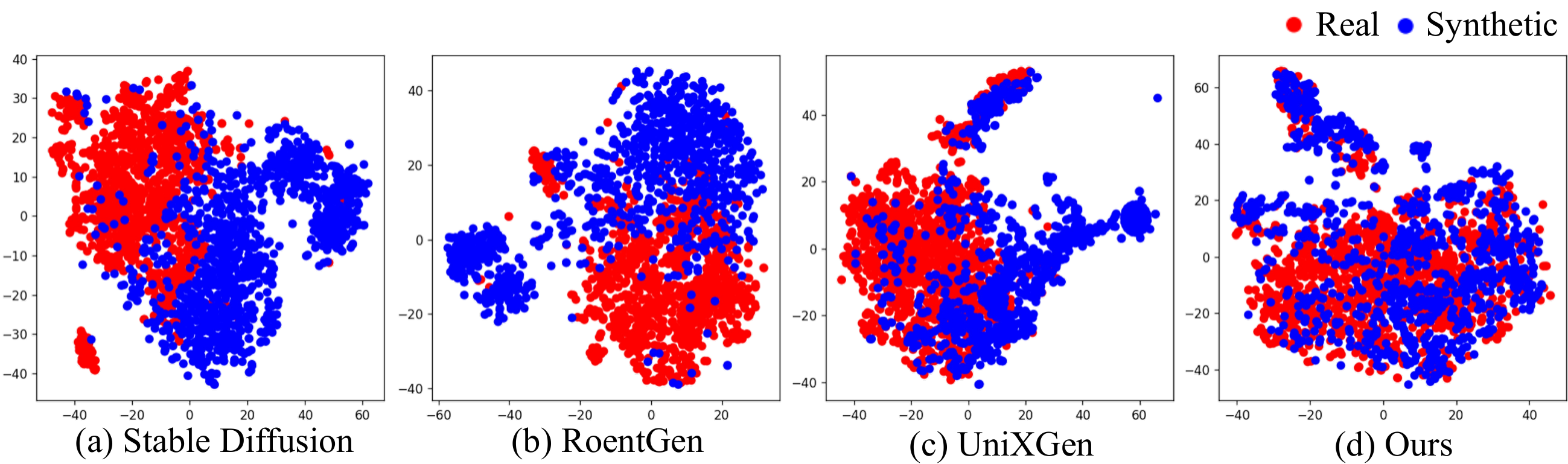}
   \caption{The t-SNE visualization of the real and synthetic CXR images on the MIMIC-CXR dataset.}
   \label{fig:tsne}
\end{figure}

\begin{table*}[t]
	\centering
	\renewcommand{\arraystretch}{1.2}
    \setlength{\tabcolsep}{0.2cm}
	\caption{Anatomy and pathology classification performance (\%) comparison of MIMIC-CXR dataset and CXR images generated by our method. }
	\begin{tabular}{c|c|c|c|c|c|c}
		\toprule
   & \multicolumn{2}{c|}{Anatomy} & \multicolumn{2}{c|}{Pathology} &  \multicolumn{2}{c}{Overall}  \\ \hline
Data source & Accuracy & AUC & Accuracy & AUC & Accuracy & AUC \\ \hline
MIMIC-CXR & 91.21 & 78.17 & \textbf{92.19} & 74.42 & 91.59 & 76.74 \\ \hline
\textbf{Ours} & \textbf{94.74} & \textbf{83.88} & 92.11 & \textbf{77.02} & \textbf{93.74} & \textbf{81.27} \\ 
		\bottomrule
	\end{tabular}
	\label{tab:ana_pat_cls}
 % \vspace{-0.1cm}
 \end{table*}

\subsection{Comparison with State-of-the-Arts}
% We quantitatively compare our method with text-to-image generation method like Stable Diffusion~\cite{rombach2022high} and report-to-CXR generation methods, such as Chambon~\etal~\cite{chambon2022adapting}, RoentGen~\cite{chambon2022roentgen}, UniXGen~\cite{lee2023unified}, and LLM-CXR~\cite{lee2023llm}. In Table~\ref{tab:img_quality}, we achieves the highest FID scores on both datasets, showing its superior effectiveness in generating CXR with descriptive reports. To compare the high-level feature distribution of CXR images generated by different methods, we randomly select 1,000 cases from test set, and apply t-SNE visualization to the real and synthetic CXR images on the MIMIC-CXR dataset. In Fig.~\ref{fig:tsne}, while current methods' synthetic CXR images differ from real ones, ours almost overlap with real ones, indicating its superiority in generating realistic CXR images.
We conducted a quantitative comparison of our method with state-of-the-art text-to-image generation methods, such as Stable Diffusion~\cite{rombach2022high}, and report-to-CXR generation approaches, including Chambon~\etal~\cite{chambon2022adapting}, RoentGen~\cite{chambon2022roentgen}, UniXGen~\cite{lee2023unified}, and LLM-CXR~\cite{lee2023llm}. As shown in Table~\ref{tab:img_quality}, our method achieves the highest FID scores on both datasets, demonstrating its superior performance in generating CXR images with descriptive reports. To further investigate the high-level feature distribution of the generated CXR images, we randomly selected 1,000 cases from the test set and performed t-SNE visualization on both real and synthetic CXR images from the MIMIC-CXR dataset. Fig.~\ref{fig:tsne} illustrates that while the synthetic CXR images generated by current methods exhibit notable differences from the real ones, our method produces images that nearly overlap with the real images in the t-SNE visualization, highlighting its exceptional ability to generate highly realistic CXR images.

% In Fig.~\ref{fig:report2cxr}, we compare CXR images generated by our method and existing methods on the MIMIC-CXR and OpenI datasets. In the first example, the proposed method accurately synthesizes `opacity near the aorta', while other methods fail to generate such features. This suggests the superior ability of our method to produce realistic CXR images based on input reports.
Fig.~\ref{fig:report2cxr} presents a comparison of CXR images generated by our method and existing approaches on both the MIMIC-CXR and OpenI datasets. In the first example, our proposed method successfully synthesizes the 'opacity near the aorta' described in the input report, while other methods struggle to generate this specific feature. This observation highlights the superior capability of our method in producing highly realistic and accurate CXR images that faithfully reflect the content of the corresponding reports.

\subsection{Semantic Analysis}
\begin{figure}[t]
  \centering
   \includegraphics[width=\linewidth]{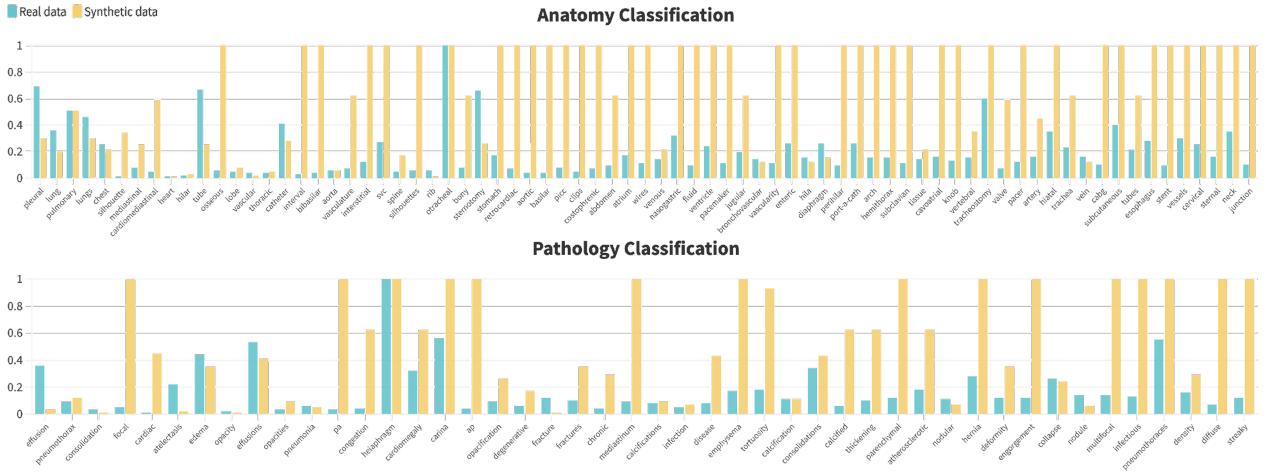}
   \caption{Anatomy and pathology classification performance of each category. Each column shows the precision score.}
   \label{fig:ana_pat_cls}
   % \vspace{-0.5cm}
\end{figure}

To further analyze the semantic information of the synthetic images, we pre-train a classifier on the MIMIC-CXR dataset for the multi-label anatomy and pathology classification. Then, we test the classification performance of the real and synthetic images. In Table~\ref{tab:ana_pat_cls}, we show the classification performance for the test set of the MIMIC-CXR dataset and CXR images generated by our method. Our method significantly outperforms the real data by a large margin with an accuracy of 2.15\%, implying our synthetic data with accurate semantic information about anatomical structures and pathological conditions. Moreover, we also show the performance of each category for anatomy and pathology classification. As visualized in Fig.~\ref{fig:ana_pat_cls}, our method achieves higher precision than the real data in most categories. These indicate the medical images generated by our method preserve more semantic information in terms of anatomy and pathology.

\section{Conclusion}
To synthesize high-quality medical images with detailed anatomical and pathology information, we introduce a medical image synthesis model to generate anatomy-pathology prompts and highly detailed medical images. In order to provide the descriptive reports with anatomy and pathology information, we design an anatomy-pathology prompting to establish anatomy and pathology vocabularies and employ GPT4 to automatically generate reports. With the descriptive reports, we devise a fine-grained alignment based synthesis module to perform alignment between the reports and pre-defined visual codebook to obtain matched keypatches. Moreover, this module utilizes the LLM and VQ-GAN to convert reports, instructions, and matched keypatches to synthetic images.
\bibliographystyle{splncs04}
\bibliography{mybibliography}
%

% \bibliography{bib}

\end{document}